\documentclass[acmtog]{acmart}

\AtBeginDocument{%
  }

\copyrightyear{2026}
\acmYear{2026}
\setcopyright{cc}
\setcctype{by}
\acmConference[SA Conference Papers '26]{SIGGRAPH Asia 2026 Conference Papers}{December 01--04, 2026}{Kuala Lumpur, Malaysia}
\acmBooktitle{SIGGRAPH Asia 2026 Conference Papers (SA Conference Papers '26), December 01--04, 2026, Kuala Lumpur, Malaysia}
\acmDOI{10.1145/3829340.3842320}
\acmISBN{979-8-4007-2842-6/2026/12}

\usepackage{multirow}
\usepackage{xspace}

\newcommand{\methodname}{Artic-O\xspace}
\newcommand{\geolatent}{\mathbf{z}_{\mathrm{geo}}\xspace}
\newcommand{\imgtoken}{\mathbf{f}_{\mathrm{img}}\xspace}

\begin{document}

\title{Artic-O: End-to-End Articulated Object Reconstruction via Latent Geometry Learning}


\author{Xuyang Wang}
\authornote{Work was done during the author's research internship at KAUST.}
\affiliation{%
  \institution{Australian National University}
  \city{Canberra}
  \country{Australia}
}
\affiliation{%
  \institution{King Abdullah University of Science and Technology}
  \city{Thuwal}
  \country{Saudi Arabia}
}
\email{xuyang.wang@anu.edu.au}

\author{Zhenyu Li}
\email{zhenyu.li.1@kaust.edu.sa}
\affiliation{%
  \institution{King Abdullah University of Science and Technology}
  \city{Thuwal}
  \country{Saudi Arabia}
}

\author{Jian Ding}
\authornote{Project lead.}
\email{jian.ding@kaust.edu.sa}
\affiliation{%
  \institution{King Abdullah University of Science and Technology}
  \city{Thuwal}
  \country{Saudi Arabia}
}

\author{Habib Slim}
\email{habib.slim@kaust.edu.sa}
\affiliation{%
  \institution{King Abdullah University of Science and Technology}
  \city{Thuwal}
  \country{Saudi Arabia}
}

\author{Peter Wonka}
\email{peter.wonka@kaust.edu.sa}
\affiliation{%
  \institution{King Abdullah University of Science and Technology}
  \city{Thuwal}
  \country{Saudi Arabia}
}

\author{Hongdong Li}
\email{hongdong.li@anu.edu.au}
\affiliation{%
  \institution{Australian National University}
  \city{Canberra}
  \country{Australia}
}

\author{Mohamed Elhoseiny}
\email{mohamed.elhoseiny@kaust.edu.sa}
\affiliation{%
  \institution{King Abdullah University of Science and Technology}
  \city{Thuwal}
  \country{Saudi Arabia}
}

\renewcommand{\shortauthors}{Xuyang Wang, Zhenyu Li, Jian Ding, Habib Slim, Peter Wonka, Hongdong Li and Mohamed Elhoseiny}

\begin{abstract}
Reconstructing articulated objects from sparse images requires recovering complete geometry, movable parts, and motion parameters. 
Recent methods typically separate geometry reconstruction, part reasoning, and articulation estimation into different stages. 
This separation can weaken consistency between shape, active parts, and motion, while also incurring substantial inference cost.
We introduce \methodname, an end-to-end, feed-forward framework for articulated object reconstruction via latent geometry learning. 
Instead of fitting geometry in image or view space, \methodname maps sparse multi-state observations into a pretrained latent geometry space, where a frozen flow-matching decoder provides a complete-shape prior for recovering visible and occluded structures. 
To connect geometry with articulation, \methodname fuses visual tokens, geometry latents, and point-wise decoder features in an image-grounded part-reasoning module for active-part segmentation and articulation prediction. 
We further train the model with a geometry-to-articulation curriculum and a decoupled two-pass strategy to balance reconstruction and part-level supervision.
On PartNet-Mobility, \methodname achieves strong reconstruction quality while being substantially more efficient than LARM, a strong prior method. It reduces Chamfer Distance, improves F-score, and achieves comparable or better articulation accuracy across most joint metrics, while reducing inference time from $9$ minutes to about $0.3$ seconds per object. Code is available at \url{https://github.com/Wxyxixixi/Artic-O}.

\end{abstract}

\begin{teaserfigure}
  \includegraphics[width=0.95\textwidth]{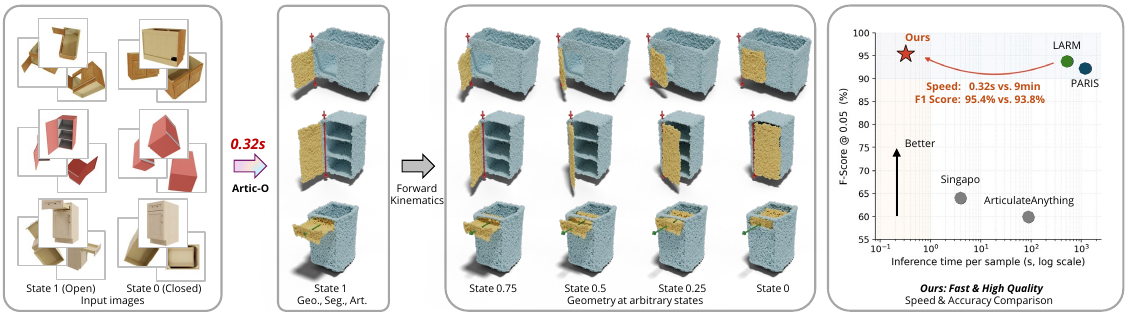}
  \caption{
  \textbf{Artic-O reconstructs articulated objects from sparse multi-state images in a feed-forward manner.}
Given only a few input views of an object captured at different articulation states, Artic-O predicts the object geometry, part segmentation, and articulation parameters.
These outputs define an articulated asset that can be animated through forward kinematics to synthesize geometry at novel articulation states.
Compared with prior optimization-based methods, Artic-O achieves strong reconstruction quality while enabling fast articulated asset reconstruction.
  }
\Description{
Pipeline overview of Artic-O. Sparse images of an articulated object at multiple motion states are shown on the left. The model reconstructs complete 3D geometry, identifies the movable and static parts, and estimates the articulation parameters. On the right, the reconstructed object is shown animated at several novel articulation states.
}
  \label{fig:teaser}
\end{teaserfigure}

\maketitle

\section{Introduction}

Articulated objects, such as cabinets, drawers, and storage boxes, are pervasive in real-world environments.  Compared with static objects, modeling articulated objects requires recovering not only their overall shape, but also the movable parts and motion mechanisms that define how they can be interacted with~\cite{wang2019shape2motion,liu2025survey}. This capability is important for creating interactive 3D assets that can support a wide range of downstream applications, such as  digital twins, AR/VR content creation, animation, physical simulation, and embodied intelligence ~\cite{xiang2020sapien,jiang2022ditto,liu2025survey}.

Existing methods for articulated object reconstruction can be broadly grouped into two categories based on input modality.
\textbf{\textit{(1) Geometry-based methods}} estimate movable parts and articulation parameters from clean point clouds or 3D shapes~\cite{fu2024capt,li2025particulate, qiu2025articulate}.
While enabling high-quality articulation reasoning directly in 3D space, these methods assume access to accurate and complete object geometry, which is often unrealistic in real-world deployments.
\textbf{\textit{(2) Image-based methods}} relax this requirement by starting from visual observations, but inferring complete articulated geometry from images is highly ambiguous.
As a result, many approaches rely on \textit{per-instance optimization}~\cite{liu2023paris,chen2025freeart3d} or \textit{retrieved parts} from a geometry library~\cite{chen2024urdformer,le2025articulate,liu2025singapo}, which can limit geometric fidelity, scalability, or adherence to the input images.

A recent method, LARM ~\cite{yuan2025larm}, takes an important step toward scalable image-based articulated object reconstruction by learning a feed-forward model that recovers object geometry from sparse multi-state observations without per-instance geometry optimization. 
A direct strategy is to reconstruct the two observed states independently and infer articulation from their geometric correspondences. LARM goes beyond such two-state fitting by synthesizing additional intermediate joint states to establish more reliable correspondences for articulation estimation.
Nevertheless, its overall pipeline still remains a separate downstream stage that depends on the reconstructed geometry and cross-state correspondences. This separation can propagate errors from reconstruction into final motion estimation, leading to suboptimal results. 

By contrast, a closely related work, ART~\cite{li2025art}, studies end-to-end articulated object reconstruction by directly predicting individual part-level geometry and articulation. Our method instead leverages latent geometry prior for complete point-cloud reconstruction and joint articulation prediction.

In this work, we propose \methodname, an \textit{end-to-end} framework for sparse multi-state articulated object reconstruction based on \textit{latent geometry learning}.
As shown in Fig.~\ref{fig:teaser}, given a few images of an articulated object captured at two endpoint states, 
\methodname~jointly reasons across states to predicts complete object geometry, active-part segmentation, and articulation parameters \textit{simultaneously} in a unified feed-forward pipeline. The core idea is to extend the \textit{geometry latent} from static object reconstruction to articulated object reconstruction.
Specifically, we leverage a pretrained flow-matching shape decoder learned from complete object point clouds and keep it frozen during training.
A trainable state-aware image encoder maps sparse multi-state observations into this latent space, allowing the decoder to recover object-level geometry beyond the directly visible surfaces. 
On top of the learned \textit{geometry latent}, \methodname performs image-grounded part and articulation reasoning.
Specifically, we distill the learned geometry together with the image features into a compact part memory that guides active-part segmentation and provides a part-level representation for articulation prediction.
The articulation head then estimates the corresponding joint parameters from this active-part representation and point-wise geometry features.
Since these components are trained together in an end-to-end manner, it turns sparse multi-state images into an articulated asset without inference-time optimization.

Training this \textit{multi-task} system requires balancing generative geometry reconstruction with discriminative part-level reasoning.
We therefore adopt a geometry-to-articulation \textit{curriculum} strategy: the model first learns to predict image-conditioned latents that can be decoded into a point cloud by the frozen shape decoder, and then introduces segmentation and articulation supervision while preserving the decoder as a stable geometry prior.
We further use a decoupled two-pass training strategy: one decoder pass optimizes flow matching over noisy trajectory samples, while the other trains segmentation and articulation on a mixture of clean point cloud and noisy point cloud.
Clean queries provide reliable part identity, whereas noisy queries prevent the heads from solely relying on perfect ground-truth geometry.
This mixed query distribution closely matches the imperfect reconstructed shapes observed at inference time, improving the robustness of part and articulation prediction.

Experiments on PartNet-Mobility~\cite{xiang2020sapien} show that \methodname improves sparse articulated reconstruction over LARM.
It reduces Chamfer Distance from $0.019$ to $0.017$ and improves F-score from $93.8\%$ to $95.4\%$, while achieving comparable or better articulation accuracy across most joint metrics, including improving motion-range accuracy from $87.0\%$ to $93.5\%$.
Our feed-forward design also makes inference substantially more efficient, taking $0.32$ seconds per object compared to $9$ minutes required by LARM.
Ablations further confirm the importance of the frozen geometry prior, image-grounded part reasoning, curriculum training, and the decoupled two-pass strategy. We additionally demonstrate an extension to multi-part objects through independent part-wise inference.

\section{Related Work}
\label{sec:related}

\subsection{Static Object Generation}
Image-to-3D methods have made substantial progress in reconstructing and generating complete static objects from limited observations. 
Large reconstruction models~\cite{hong2024lrm, xu2024grm, wei2024meshlrm} formulate single-image 3D reconstruction as feed-forward prediction, while subsequent methods improve geometry quality, texture fidelity, and multi-view consistency with stronger image conditioning and expressive 3D representations~\cite{TripoSR2024,tang2024lgm, yang2024hunyuan3d,zhao2025hunyuan3d,li2025triposg,xiang2025structured, zhang2024clay}. 
These works demonstrate that learned object-level priors can infer plausible complete geometry from sparse or incomplete image evidence. 
Recent part-aware methods further decompose static 3D objects into semantic or structural parts~\cite{yang2024sampart3d, zhou2025pointsam, zhu2026partsam,yan2025x, lin2026partcrafter, yang2025omnipart}. 
However, their targets remain \textit{static}: they do not infer movable parts and articulation, such as motion axes and ranges. 
Moreover, PhysX-3D~\cite{cao2025physxd} and PhysX-Anything~\cite{cao2026physx} extend image-to-3D generation toward physically grounded and interactive assets, incorporating kinematic and physical properties into the generated 3D objects. By contrast, our goal is to reconstruct articulated objects from sparse multi-state observations.

\subsection{Articulated Object Reconstruction}
Geometry-based methods infer articulation from explicit 3D input, such as complete or partial point clouds, depth scans, part-segmented point clouds, and meshes, estimating movable parts, part poses, canonical coordinates, or motion attributes from direct geometric observations~\cite{wang2019shape2motion,yan2019rpm,li2020category,weng2021captra,fu2024capt,goyal2025geopard,li2025particulate,le2025articulate}. Although these methods have advanced part-level articulation estimation, their reliance on reliable 3D geometry limits their applicability when only sparse images are available.

Image-conditioned methods remove the requirement of a clean 3D input, but obtain geometry through different mechanisms. 
Retrieval- or asset-based methods ~\cite{chen2024urdformer, liu2025singapo, le2025articulate} infer articulation from images or multimodal prompts, but their geometry is obtained through retrieved or assembled assets rather than direct reconstruction of the observation. Optimization-based methods~\cite{liu2023paris,chen2025freeart3d} instead recover detailed geometry by fitting or refining object-specific representations from visual observations, while remain time-consuming. 

Sparse multi-state reconstruction methods~\cite{yuan2025larm, zhao2025real2code} further exploit images captured under different articulation states to provide complementary cues for geometry completion and movable-part reasoning, but rely on pipeline with separate stages. 
Concurrent work ART~\cite{li2025art} also studies end-to-end articulated object reconstruction from sparse multi-state images, but adopts a part-based representation that jointly decodes per-part geometry, texture, and articulation from learnable part slots.
\methodname instead operates on an object-level geometry latent, using a pretrained shape decoder for complete geometry reconstruction and the resulting geometry features for part and articulation reasoning.

\subsection{Geometry Latent Learning}
Geometry latent learning provides a representation-level foundation for complete 3D generation and reconstruction. 
Instead of predicting geometry only as pixel-aligned depth, point maps, or per-view fields, these methods encode 3D shapes into compact latent variables, latent sets, or structured tokens that can be decoded into complete geometry. 
3DShape2VecSet~\cite{zhang20233dshape2vecset} introduces a set-based latent representation for neural-field generation, while TRELLIS~\cite{xiang2025structured} and TripoSG~\cite{li2025triposg} demonstrate that structured 3D latents and large-scale generative models can support high-quality image-conditioned 3D asset generation. Additionally, FreeArt3D~\cite{chen2025freeart3d} leverages the TRELLIS prior to guide per-instance optimization for articulated reconstruction.
NOVA3R~\cite{chennova3r} further studies non-pixel-aligned reconstruction from unposed images, aggregating image evidence into global latent tokens and decoding complete point clouds instead of per-view surface layers, enabling object-level point-cloud completion with improved geometry quality and multi-view consistency.
Building on this principle, our method extends geometry latent learning from static shape reconstruction to articulated reconstruction, where geometry latents serve not only as a complete-shape prior but also as structural context for image-grounded active-part segmentation and articulation prediction.

\section{Method}
\label{sec:method}

\begin{figure*}[t!]
    \centering
    \includegraphics[width=0.95\linewidth]{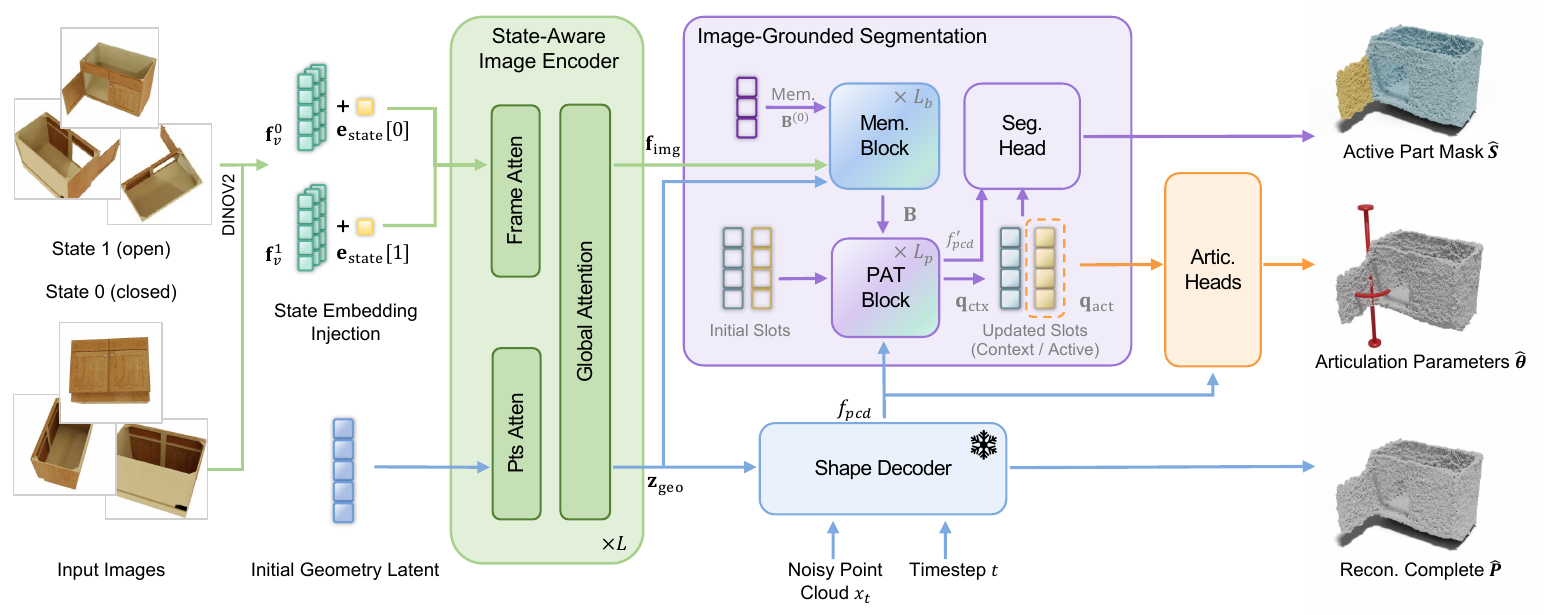}
    \caption{
\textbf{Overview of \methodname.}
Given sparse two-state images, \methodname predicts geometry latents $\mathbf{z}_{\mathrm{geo}}$ and image tokens $\mathbf{f}_{\mathrm{img}}$ with a \textcolor[HTML]{58B39C}{\textbf{State-Aware Image Encoder}}.
A frozen flow-matching based \textcolor[HTML]{6F9FDC}{\textbf{Shape Decoder}} reconstructs the complete point cloud $\hat{\mathbf{P}}$, while \textcolor[HTML]{9B75D6}{\textbf{Image-Grounded Segmentation}} and slot-driven \textcolor[HTML]{F57C20}{\textbf{Articulation Heads}} predict the active-part mask $\hat{\mathbf{S}}$ and articulation parameters $\hat{\boldsymbol{\Theta}}$.
The whole framework is trained in an end-to-end manner.
    }
\Description{
Method overview of Artic-O. Multi-viewed multi-state images inputs to the model through a state-aware image encoder. Initial geometry latent are given to the point couds attention module. The image feature and the geometry latent will cross-attention with each others. Then the Image grounded segmentation module and the shape decoder will decode the input latents into segmentation labels and point cloud geometry. A pair of part slot tokens will be given to the articulation heads to predicts the articulation. 
}
\label{fig:overview}
\end{figure*}

\subsection{Method Overview}

\paragraph{Task formulation.}
We adopt the same sparse two-state observation setting as LARM~\cite{yuan2025larm}, where an articulated object is observed by a few views at two different articulation states.
In contrast to LARM which formulates the task as articulation-conditioned novel-view synthesis followed by reconstruction and optimization for articulation recovery, \methodname directly predicts object-level 3D outputs in a feed-forward manner.
Given the sparse multi-state image set $\mathcal{I}$, \methodname predicts a 3D point cloud $\hat{\mathbf{P}}\in\mathbb{R}^{N\times 3}$, a binary active-part segmentation $\hat{\mathbf{S}}\in\{0,1\}^{N}$, and articulation parameters $\hat{\boldsymbol{\Theta}}$:
$
\hat{\mathbf{P}}, \hat{\mathbf{S}}, \hat{\boldsymbol{\Theta}}
=
\text{\methodname}(\mathcal{I}).
$
The input image set consists of images captured at two endpoint articulation states,
$
\mathcal{I}=(\mathbf{I}^{0},\mathbf{I}^{1})
$
with
$
\mathbf{I}^{s}=\{I_v^s\}_{v=1}^{V},
$
where $s\in\{0,1\}$ indexes the articulation state and $v$ indexes the camera view.
Following LARM~\cite{yuan2025larm}, we use $V=3$ views per state by default, resulting in six input images per object.

\paragraph{Method overview.}
\methodname devises a feed-forward pipeline that first maps sparse multi-state images into a pretrained latent geometry space, then decodes geometry, segmentation, and articulation from this representation.
As shown in Fig.~\ref{fig:overview}, a state-aware image encoder takes $\mathcal{I}$ as input and predicts image-conditioned geometry latents $\geolatent$ together with image tokens $\imgtoken$ (Sec.~\ref{sec:method:image_encoder}).
The geometry latents lie in the latent space of a pretrained point-cloud autoencoder~\cite{chennova3r}, whose frozen flow-matching shape decoder reconstructs the point cloud $\hat{\mathbf{P}}$ and provides point-wise geometry features $\mathbf{f}_{\mathrm{pcd}}$ (Sec.~\ref{sec:method:geo_prior}).
To identify the moving part, an image-grounded segmentation head combines $\mathbf{f}_{\mathrm{pcd}}$, $\geolatent$, and $\imgtoken$ to predict the active-part mask $\hat{\mathbf{S}}$ (Sec.~\ref{sec:method:seg_head}).
Finally, a slot-driven articulation head uses the active-part representation and point-wise geometry features $\mathbf{f}_{\mathrm{pcd}}$ to estimate the articulation parameters $\hat{\boldsymbol{\Theta}}$ for revolute or prismatic motion (Sec.~\ref{sec:method:artic_head}).

\begin{figure}[t!]
    \centering
    \includegraphics[width=0.85\linewidth]{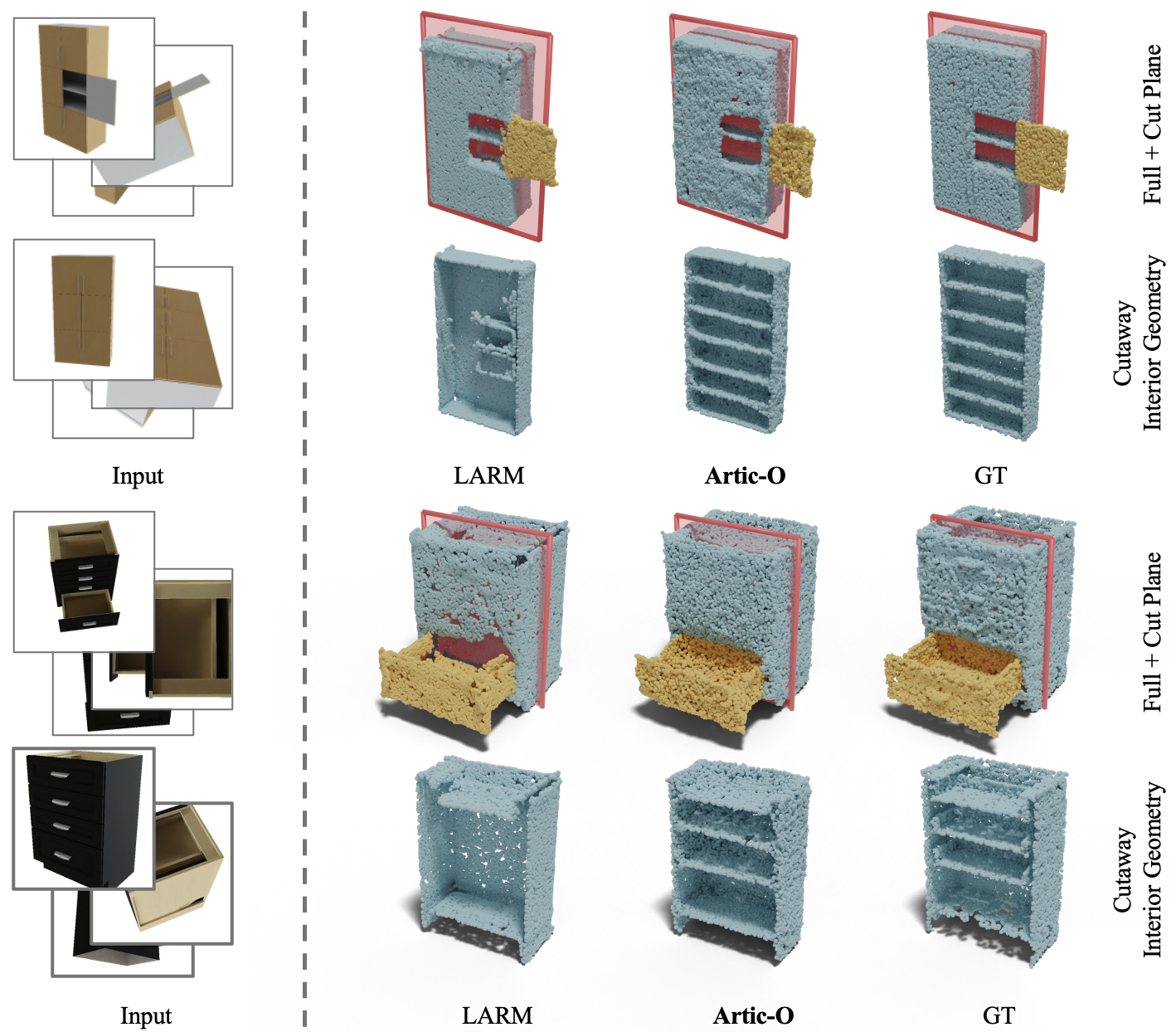}
    \caption{
\textbf{Cutaway comparison.}
We show the interior geometry of two reconstructed objects from our test set. The red plane indicates the cut location, and the cutaway view removes the camera-facing half of the reconstructed point cloud.
Compared with LARM, \methodname recovers a fuller interior volume that is closer to the GT, illustrating the benefit of the latent geometry prior.}
\Description{
Showcase of the interior geometry reconstruction results of the LARM, Artic-O and GT. On the left, shows the input sparse view. On the rights shows the recontruction result with a cut away plane.
}

    \label{fig:cut}
\end{figure}

\subsection{Latent Geometry Autoencoder}
\label{sec:method:geo_prior}

\methodname builds upon a latent point-cloud autoencoder~\cite{chennova3r}.
Given a complete point cloud $\mathbf{P}\in\mathbb{R}^{N\times 3}$, the point-cloud encoder maps it to a fixed-size set of latent geometry tokens
$
\mathbf{z}_{\mathrm{geo}}
=
\mathcal{E}_{\mathrm{geo}}(\mathbf{P})
\in\mathbb{R}^{K\times d}.
$
Conditioned on these tokens, a flow-matching (FM) shape decoder predicts a velocity field for noisy point queries:

\begin{equation}
    \mathbf{v}_t
    =
    \mathcal{D}_{\mathrm{geo}}(\mathbf{x}_t,t;\mathbf{z}_{\mathrm{geo}}),
\end{equation}
where $\mathbf{x}_t$ is the noisy point query at flow timestep $t$ and $\mathbf{v}_t$ is the predicted velocity along the flow trajectory.
Starting from sampled noise, the decoder reconstructs a point cloud by integrating the predicted velocity field.

In \methodname, we only use two components of this design. The pre-trained latent space itself (encoding a shape as a fixed-size set of latent tokens $\geolatent$) and the 
 shape decoder $\mathcal{D}_{\mathrm{geo}}$ which we keep frozen.
The point-cloud encoder $\mathcal{E}_{\mathrm{geo}}$ is not used.

Fig.~\ref{fig:cut} visualizes the advantage of using this latent space: the latent geometry prior encodes interior geometry, whereas the view-conditioned reconstruction of LARM tends to leave hollow regions.

\subsection{State-Aware Image Encoder}
\label{sec:method:image_encoder}

We modify the image encoder of NOVA3R~\cite{chennova3r}, an improvement of VGGT encoder~\cite{wang2025vggt}, to the sparse two-state articulated setting.
Given the images $\mathcal{I}$ and learnable geometry tokens $\mathbf{Q}^{(0)}$ as queries, the encoder outputs geometry latents and image tokens:
\begin{equation}
    \imgtoken, \geolatent
    =
    \mathcal{E}_{\mathrm{img}}(\mathcal{I},\mathbf{Q}^{(0)}),
\end{equation}
where $\imgtoken$ provides multi-view visual information for segmentation and articulation, while $\geolatent$ conditions the frozen shape decoder and is also reused by downstream heads.

The main adaptation is a lightweight state-aware token embedding.
For each image $I_v^s$ from articulation state $s\in\{0,1\}$, we extract image patch tokens $\mathbf{f}_v^s\in\mathbb{R}^{P\times C}$ using a shared image backbone~\cite{oquab2023dinov2}.
We introduce a learnable state embedding table $\mathbf{e}_{\mathrm{state}}\in\mathbb{R}^{2\times C}$ and add the corresponding embedding to all patch tokens from the same state:
\begin{equation}
    \tilde{\mathbf{f}}_v^s
    =
    \mathbf{f}_v^s
    +
    \mathbf{e}_{\mathrm{state}}[s].
\end{equation}
The state-aware image tokens from all views and states are then processed together with the learnable geometry tokens by the VGGT-style transformer~\cite{wang2025vggt}.
This keeps the backbone architecture unchanged while allowing the encoder to distinguish the two observed articulation states. The encoder is trained to encode only the geometry of state 0.

\subsection{Image-Grounded Active-Part Segmentation}
\label{sec:method:seg_head}

To extend the framework from static shape to articulated object, we predict which reconstructed points belong to the active moving part.
The segmentation head takes point-wise geometry features $\mathbf{f}_{\mathrm{pcd}}\in\mathbb{R}^{N\times d}$ from the frozen flow-matching decoder, image tokens $\imgtoken$, and geometry latents $\geolatent$ as input.
The point-wise features provide local 3D geometry, while the image and latent tokens provide multi-view, multi-state evidence for separating the active part from the static context.

\paragraph{Image-grounded part memory.}
Directly using all image tokens for point-wise segmentation is inefficient, since motion cues are scattered across views, patches, and states.
We therefore introduce an image-grounded part memory that distills dense image tokens, together with geometry latents, into a compact set of part-oriented tokens.

Concretely, the part memory is initialized as learnable query tokens $\mathbf{B}^{(0)}\in\mathbb{R}^{M\times d}$.
We update the memory tokens by cross-attention with image tokens $\mathbf{f}_{\mathrm{img}}$ and projected geometry latents $\geolatent$:
\begin{equation}
    \mathbf{B}^{(\ell+1)}
    =
    \mathrm{MemBlock}^{(\ell)}
    \left(
        \mathbf{B}^{(\ell)},\mathbf{f}_{\mathrm{img}}, \geolatent
    \right),
    \ell=0,\dots,L_b-1.
\end{equation}
The resulting memory $\mathbf{B}$ provides a compact image-grounded summary for active-part segmentation.

\paragraph{Memory-grounded slot segmentation.}
Given the part memory, we perform binary segmentation with two fixed-role slots,
$
\mathbf{q}_{\mathrm{ctx}}
$
for the static context and
$
\mathbf{q}_{\mathrm{act}}
$
for the active moving part.
Since each object contains at most one active part in our setting, the slot semantics are fixed and no permutation matching is required.

Starting from $\mathbf{x}^{(0)}=\mathbf{f}_{\mathrm{pcd}}$, we refine the slots and point features with memory-grounded part-attention blocks:
\begin{equation}
    \mathbf{Q}_{\mathrm{seg}}^{(\ell+1)},\mathbf{x}^{(\ell+1)}
    =
    \mathrm{PATBlock}^{(\ell)}
    \left(
        \mathbf{Q}_{\mathrm{seg}}^{(\ell)},\mathbf{x}^{(\ell)},\mathbf{B}
    \right),
    \ell=0,\dots,L_p-1.
\end{equation}
Each block lets the slots aggregate part-level evidence from the memory and point features, and then propagates the updated slot information back to the point features.
After $L_p$ blocks, a shared MLP predicts point-slot logits
$
\ell_{i,c}
=
\mathrm{MLP}_{\mathrm{seg}}([\mathbf{f}'_{\mathrm{pcd}};\mathbf{q}_c])
$
for
$
c\in\{\mathrm{ctx},\mathrm{act}\},
$
where $\mathbf{f}'_{\mathrm{pcd}}=\mathbf{x}^{L_p-1}$.
The active-part mask is obtained by
$
\hat{S}_i=\arg\max_c \ell_{i,c}.
$
The final active slot $\mathbf{q}_{\mathrm{act}}$ is used as the part-level representation for articulation prediction.

\subsection{Slot-Driven Articulation Prediction}
\label{sec:method:artic_head}

After locating an active part, \methodname predicts its articulation parameters from the active slot $\mathbf{q}_{\mathrm{act}}$ and the point-wise geometry features $\mathbf{f}_{\mathrm{pcd}}$. Following~\cite{yuan2025larm, li2025particulate}, we consider two common types of part motion: prismatic and revolute. Prismatic motion corresponds to linear translation, such as a drawer sliding in and out, and is parameterized by a direction axis $\mathbf{a}_{\mathrm{pri}}$ and a translation range $r_{\mathrm{pri}}$. Revolute motion corresponds to rotation around the axis, such as door opening or closing, and is parameterized by a rotation axis $\mathbf{a}_{\mathrm{rev}}$ and an angular range $r_{\mathrm{rev}}$.
The active slot captures part-level information, while the point-wise features preserve local geometry for axis localization.
We use lightweight MLP heads to predict revolute and prismatic parameters from $\mathbf{q}_{\mathrm{act}}$:
\begin{equation}
    \hat{\mathbf{a}}_{\mathrm{rev}}
    =
    h_{\mathrm{rev}}^{\mathrm{dir}}(\mathbf{q}_{\mathrm{act}}),
    \quad
    \hat{r}_{\mathrm{rev}}
    =
    \mathrm{softplus}
    \left(
        h_{\mathrm{rev}}^{\mathrm{range}}(\mathbf{q}_{\mathrm{act}})
    \right),
\end{equation}
\begin{equation}
    \hat{\mathbf{a}}_{\mathrm{pri}}
    =
    h_{\mathrm{pri}}^{\mathrm{dir}}(\mathbf{q}_{\mathrm{act}}),
    \quad
    \hat{r}_{\mathrm{pri}}
    =
    \mathrm{softplus}
    \left(
        h_{\mathrm{pri}}^{\mathrm{range}}(\mathbf{q}_{\mathrm{act}})
    \right).
\end{equation}
The predicted directions are normalized at inference time.
The motion type is provided by object annotation, and the corresponding branch is used for supervision and evaluation.

\paragraph{Dense revolute-axis voting.}
For revolute joints, we follow PARTICULATE~\cite{li2025particulate}: the active slot predicts the axis direction $\mathbf{l}_{\mathrm{rev}}$, while each point votes for the axis location $\mathbf{c}_i$ by predicting its closest point on the revolute axis.
For each point $i$, we predict
$
\hat{\mathbf{c}}_i
=
h_{\mathrm{cp}}([\mathbf{f}_{\mathrm{pcd},i};\mathbf{q}_{\mathrm{act}}])
\in\mathbb{R}^{3}.
$
At inference time, we normalize
$
\hat{\mathbf{l}}_{\mathrm{rev}}
=
\hat{\mathbf{a}}_{\mathrm{rev}}/\|\hat{\mathbf{a}}_{\mathrm{rev}}\|_2
$
and aggregate closest-point votes over predicted active points:
$
\hat{\mathbf{o}}_{\mathrm{rev}}
=
\mathrm{Median}(\{\hat{\mathbf{c}}_i\mid \hat{S}_i=1\}).
$
The final revolute axis is assembled as the Pl\"ucker line
$
\hat{\boldsymbol{\pi}}_{\mathrm{rev}}
=
[
\hat{\mathbf{l}}_{\mathrm{rev}};
\hat{\mathbf{l}}_{\mathrm{rev}}\times\hat{\mathbf{o}}_{\mathrm{rev}}
].
$

\begin{figure}[t]
    \centering
    \includegraphics[width=0.8\linewidth]{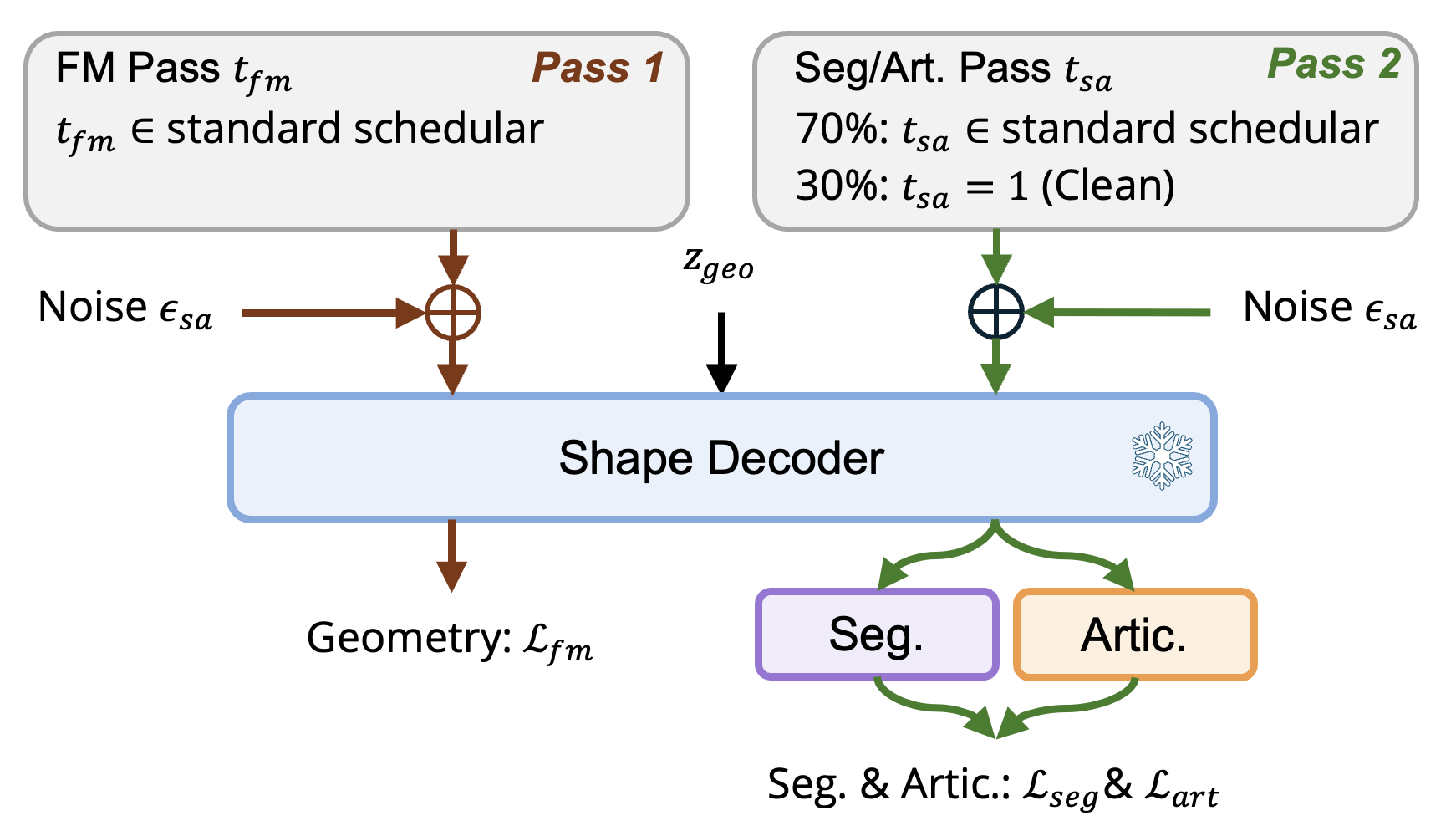}
    \caption{
\textbf{Decoupled two-pass training.}
We share one encoder but run the frozen shape decoder twice: an FM pass with the standard noise schedule for $\mathcal{L}_{\mathrm{fm}}$, and a segmentation/articulation pass biased toward clean geometry for $\mathcal{L}_{\mathrm{seg}}$ and $\mathcal{L}_{\mathrm{art}}$.
}
\Description{
Two-pass training scheme diagram overview. 
}
    \label{fig:twopass}
\end{figure}

\subsection{Training Strategy}
\label{sec:method:training}

\paragraph{Training losses.}
We train \methodname with a multi-task objective:
$
\mathcal{L}
=
\mathcal{L}_{\mathrm{fm}}
+
\lambda_{\mathrm{seg}}\mathcal{L}_{\mathrm{seg}}
+
\mathcal{L}_{\mathrm{art}}.
$
We use superscript $*$ to denote ground-truth. The frozen shape decoder is employed for computing the flow-matching loss, i.e., for sampled noise $\boldsymbol{\epsilon}$, target point cloud $\mathbf{P}^{*}$, and
$
\mathbf{x}_t=(1-t)\boldsymbol{\epsilon}+t\mathbf{P}^{*},
$
the decoder predicts
$
\hat{\mathbf{v}}_t=\mathcal{D}_{\mathrm{geo}}(\mathbf{x}_t,t;\geolatent)
$
and is supervised by
$
\mathcal{L}_{\mathrm{fm}}
=
\|\hat{\mathbf{v}}_t-(\mathbf{P}^{*}-\boldsymbol{\epsilon})\|_2^2,
$ where $t\in[0, 1]$ {\emph{per}} the standard flow-matching schedule.
The segmentation loss is
$
\mathcal{L}_{\mathrm{seg}}
=
\operatorname{CE}(\hat{\mathbf{S}},\mathbf{S}^{*}),
$
where $\mathbf{S}^{*}$ is the binary context/active label.

The articulation loss is
$
\mathcal{L}_{\mathrm{art}}
=
\mathcal{L}_{\mathrm{rev}}
+
\mathcal{L}_{\mathrm{pri}},
$
with revolute terms applied only to revolute samples and prismatic terms only to prismatic samples.
For revolute joints,
$
\mathcal{L}_{\mathrm{rev}}
=
\lambda_{\mathrm{rev\_dir}}\|\hat{\mathbf{a}}_{\mathrm{rev}}-\mathbf{a}^{*}_{\mathrm{rev}}\|_1
+
\lambda_{\mathrm{rev\_range}}\|\hat{r}_{\mathrm{rev}}-r^{*}_{\mathrm{rev}}\|_1
+
\lambda_{\mathrm{rev\_cp}}\|\hat{\mathbf{c}}-\mathbf{c}^{*}\|_1
+
\lambda_{\mathrm{unit}}|\|\hat{\mathbf{a}}_{\mathrm{rev}}\|_2-1|,
$
where $\mathbf{c}^{*}$ is the analytical closest point on the ground-truth revolute axis and the closest-point loss is computed only on ground-truth active points.
For prismatic joints,
$
\mathcal{L}_{\mathrm{pri}}
=
\lambda_{\mathrm{pri\_dir}}\|\hat{\mathbf{a}}_{\mathrm{pri}}-\mathbf{a}^{*}_{\mathrm{pri}}\|_1
+
\lambda_{\mathrm{pri\_range}}\|\hat{r}_{\mathrm{pri}}-r^{*}_{\mathrm{pri}}\|_1
+
\lambda_{\mathrm{unit}}|\|\hat{\mathbf{a}}_{\mathrm{pri}}\|_2-1|.
$
The range heads use softplus activations to enforce non-negative motion ranges.

\paragraph{Geometry-to-articulation curriculum.}
Joint training of geometry reconstruction, active-part segmentation, and articulation prediction may lead to conflicting gradients on the shared image encoder. We therefore adopt a two-stage curriculum learning scheme. Specifically, in stage-1 we optimize only $\mathcal{L}_{\mathrm{fm}}$ on articulated objects, keeping the pretrained flow-matching decoder frozen and updating only the trainable encoder-side components.
This adapts the image-conditioned latent representation to the two-state articulated setting before applying any segmentation or articulation supervision.
In stage-2, we initialize from this checkpoint, keep the decoder frozen, and train the segmentation and articulation heads together with the image encoder.
Here, the segmentation and articulation heads are optimized with a learning rate $10\times$ larger than the image encoder, allowing the heads to learn quickly while limiting drift of the geometry representation.
This allows the encoder to adapt to part-level supervision while preserving a stable geometry prior.

\paragraph{Decoupled two-pass training.}
Within the above stage-2, we use a two-pass schedule for each training step, as in Fig.~\ref{fig:twopass}.
The flow-matching loss benefits from noisy point queries sampled along the full flow trajectory, whereas segmentation and articulation require point-wise features with meaningful part identity.
We therefore share one encoder forward but run the frozen decoder twice.
In the first pass, timestep $t_{\mathrm{fm}}\in[0, 1]$ is sampled from the standard flow-matching schedule, and we optimize $\mathcal{L}_{\mathrm{fm}}$ (to optimize the image encoder, geometry decoder remains frozen).
In the second pass, we sample an independent noise point set $\boldsymbol{\epsilon}_{\mathrm{sa}}$ and a separate timestep $t_{\mathrm{sa}}\in[0, 1]$ for segmentation and articulation.
In our implementation, $30\%$ of batch samples use the clean anchor $t_{\mathrm{sa}}=1$, so the query points equal $\mathbf{P}^{*}$, while the remaining $70\%$ draw $t_{\mathrm{sa}}$ from the standard noisy schedule and use
$
\mathbf{x}_{t_{\mathrm{sa}}}
=
(1-t_{\mathrm{sa}})\boldsymbol{\epsilon}_{\mathrm{sa}}
+
t_{\mathrm{sa}}\mathbf{P}^{*}.
$
The clean anchors provide unambiguous part identity, while the noisy queries expose the heads to imperfect geometry closer to the reconstructed point clouds encountered at inference.
The segmentation and articulation heads consume the point-wise decoder features from this second pass and optimize $\mathcal{L}_{\mathrm{seg}}$ and $\mathcal{L}_{\mathrm{art}}$.
At inference time, \methodname does not require test-time optimization: the encoder runs once, the reconstruction is obtained by integrating the frozen shape decoder's velocity field from sampled noise, and the segmentation and articulation heads are evaluated once on the reconstructed point cloud.
\section{Experiments}

\begin{table*}[t]
    \centering
    \caption{
    \textbf{Main comparison on PartNet-Mobility.}
All competing methods are reproduced following the LARM protocol and evaluated under the same split and metric definitions.
Following LARM~\cite{yuan2025larm}, all resulting 3D outputs are centered and normalized consistently with the PartNet-Mobility object meshes for aligned and fair metric comparison.
All speed measurements were conducted on a single NVIDIA H200 GPU.
$^\dag$: Retrieval-based methods.
    }
    \label{tab:main}
    \scalebox{0.8}{
    \begin{tabular}{p{4.6cm}|p{2.3cm}|p{1.2cm}p{1.2cm}|p{1.0cm}p{1.0cm}p{1.0cm}p{1.0cm}|p{1.5cm}|p{2.2cm}}
    \toprule
     \multirow{2}{*}{Method} & \multirow{2}{*}{Input} & \multicolumn{2}{c|}{Geometry} & \multicolumn{4}{c|}{Joint Estimation} & \multirow{2}{*}{Speed} &  \multirow{2}{*}{Ref.} \\
     &  & CD$\downarrow$ & F-Score$\uparrow$ & Angle$\uparrow$ & Origin$\uparrow$ & $M_r\uparrow$ & $M_d\uparrow$ & \\
    \midrule
    ArticulateAnything$^\dag$~\cite{le2025articulate} & single view &  0.068 & 59.8\% & 76.1\% & 77.2\% & 32.8\% & 47.8\% & $\sim$90.0s & ICLR 2025 \\
    Singapo$^\dag$~\cite{liu2025singapo} & single view & 0.062 & 63.9\% & 89.5\% & 95.9\% & 35.2\% & 57.5\% & \underline{$\sim$4.0s} & ICLR 2025 \\
    Paris~\cite{liu2023paris} & dense view & 0.029 & 92.2\% & 24.0\% & 69.9\% & 42.0\% & 42.0\% & $\sim$21min & ICCV 2023 \\
    LARM~\cite{yuan2025larm}  & sparse view & \underline{0.019} & \underline{93.8\%} & \textbf{96.7\%} & \underline{97.9\%} & \underline{87.0\%} & \textbf{95.5\%} & $\sim$9min & Sig. Asia 2025 \\
    Artic-O (Ours) & sparse view & \textbf{0.017} & \textbf{95.4\%} & \underline{96.0\%} & \textbf{98.0\%} & \textbf{93.6\%} & \underline{94.4\%} & \textbf{$\sim$0.32s} & --\\
    \bottomrule
    \end{tabular}
    }
    \Description{Main comparison on PartNet-Mobility.}
\end{table*}

Following LARM~\cite{yuan2025larm}, we use the sparse two-state setting with $V=3$ views per articulation state, resulting in six input images per object.
The input images are rendered at two endpoint joint states and resized to $224\times224$ during training.
Our state-aware image encoder follows the NOVA3R-style learnable-token image encoder built on VGGT, with a DINOv2 ViT-L/14 backbone~\cite{chennova3r,wang2025vggt,oquab2023dinov2} and 16 alternating-attention transformer blocks.
A zero-initialized learnable state embedding is added to image patch tokens according to their articulation state.
The flow-matching shape decoder is initialized from the pretrained NOVA3R latent geometry prior and kept frozen throughout training.
On top of the decoder features, we attach the image-grounded active-part segmentation head with $M=128$ part-memory tokens, updated by $L_b=2$ one-way cross-attention blocks.
We further use $L_p=2$ memory-grounded part-attention blocks to refine the two fixed-role slots, $\mathbf{q}_{\mathrm{ctx}}$ and $\mathbf{q}_{\mathrm{act}}$, and the per-point features.
The active slot is then passed to the slot-driven articulation head.

\paragraph{Dataset and splits.}
We train and evaluate on PartNet-Mobility~\cite{xiang2020sapien} using the same six categories and object-level split as LARM: StorageFurniture, Microwave, Refrigerator, Safe, TrashCan, and Table. A single \methodname model is trained jointly across all six categories; we do not train category-specific models. The training and test sets are disjoint at the object level, such that all views of each test object are unseen during training.

\paragraph{Training objectives.} Training is performed in two stages.
First, we train the encoder-side components for 24 epochs using only the flow-matching loss, while keeping the shape decoder frozen.
Second, we train the full model for 36 epochs with the decoupled two-pass schedule: one pass optimizes flow matching on noisy trajectory samples, and the other supervises active-part segmentation and articulation on near-clean point queries, with $30\%$ of samples pinned at the clean target state. The configured loss weights are $\lambda_{seg}=1.0$, $\lambda_{\mathrm{rev\_dir}}=\lambda_{\mathrm{rev\_range}}=\lambda_{\mathrm{rev\_cp}}=\lambda_{\mathrm{pri\_dir}}=\lambda_{\mathrm{pri\_range}}=0.3$, $\lambda_{\mathrm{unit}}=0.05$.
We use AdamW with learning rates $3\times10^{-4}$ for the segmentation and articulation heads and $3\times10^{-5}$ for the image encoder and state embeddings.
At inference time, \methodname predicts geometry, active-part segmentation, and articulation parameters in a feed-forward manner without test-time optimization.
The point cloud is decoded by Euler integration of the frozen flow-matching shape decoder with step size $0.04$.

\paragraph{Evaluation metrics.}
To ensure direct comparability with LARM, all methods in Table~\ref{tab:main} are reproduced and evaluated under the same LARM protocol.
We report Chamfer Distance (CD) and F-score for geometry reconstruction, and success rates for four joint-estimation metrics: axis angle, axis origin, motion range distance $M_r$, and motion direction difference $M_d$.
Axis angle measures the angular error between predicted and ground-truth joint axes, axis origin measures the shortest distance between the two axes, $M_r$ measures motion-range deviation, and $M_d$ evaluates motion-direction consistency.
Following LARM, we use success thresholds of $0.25$ radians for axis angle, $0.15$ for axis origin, and $0.3$ for both $M_r$ and $M_d$.
Geometry metrics are averaged over five articulation states $s\in\{0,0.25,0.5,0.75,1.0\}$, and joint metrics are computed on objects with non-trivial joints.

\subsection{Main Results}
\label{sec:exp:main}

\begin{figure}
    \centering
    \includegraphics[width=0.9\linewidth]{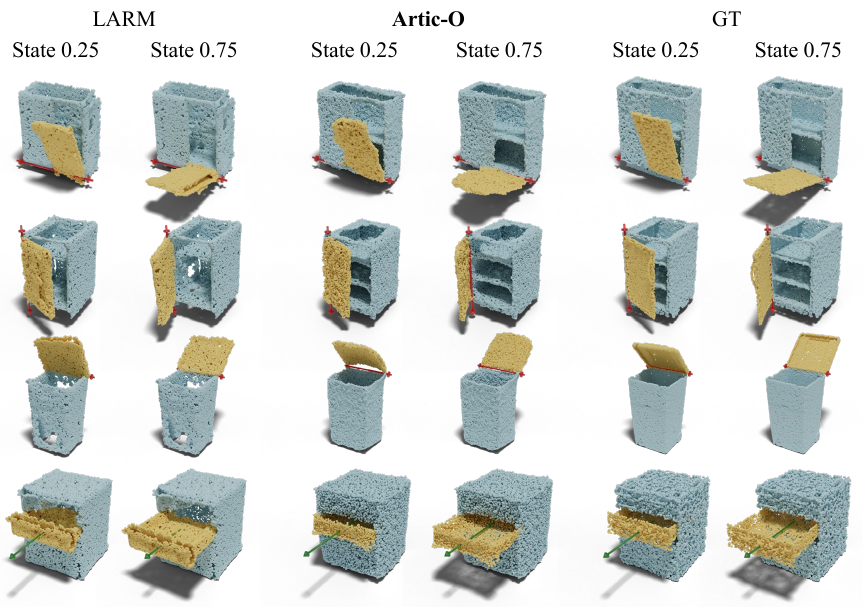}
    \caption{
\textbf{Qualitative comparison of reconstructed point clouds.}
Compared with LARM, \methodname produces more complete and coherent geometry, especially around articulated parts and sparsely observed regions.
}
\Description{
Qualitative comparison of reconstructed point clouds.
}
    \label{fig:visual_compare}
\end{figure}

Table~\ref{tab:main} compares \methodname with prior methods on PartNet-Mobility under the LARM evaluation protocol.
Compared with LARM, \methodname improves geometry reconstruction, reducing CD from $0.019$ to $0.017$ and increasing F-score from $93.8\%$ to $95.4\%$.
For joint estimation, \methodname achieves comparable or better performance across the four metrics, improving axis-origin success from $97.9\%$ to $98.0\%$ and motion-range accuracy $M_r$ from $87.0\%$ to $93.5\%$, while remaining close to LARM on axis angle and $M_d$.
These results show that \methodname improves reconstruction quality while preserving accurate articulation estimation in a feed-forward manner.

\begin{figure*}
    \centering
    \includegraphics[width=0.85\linewidth]{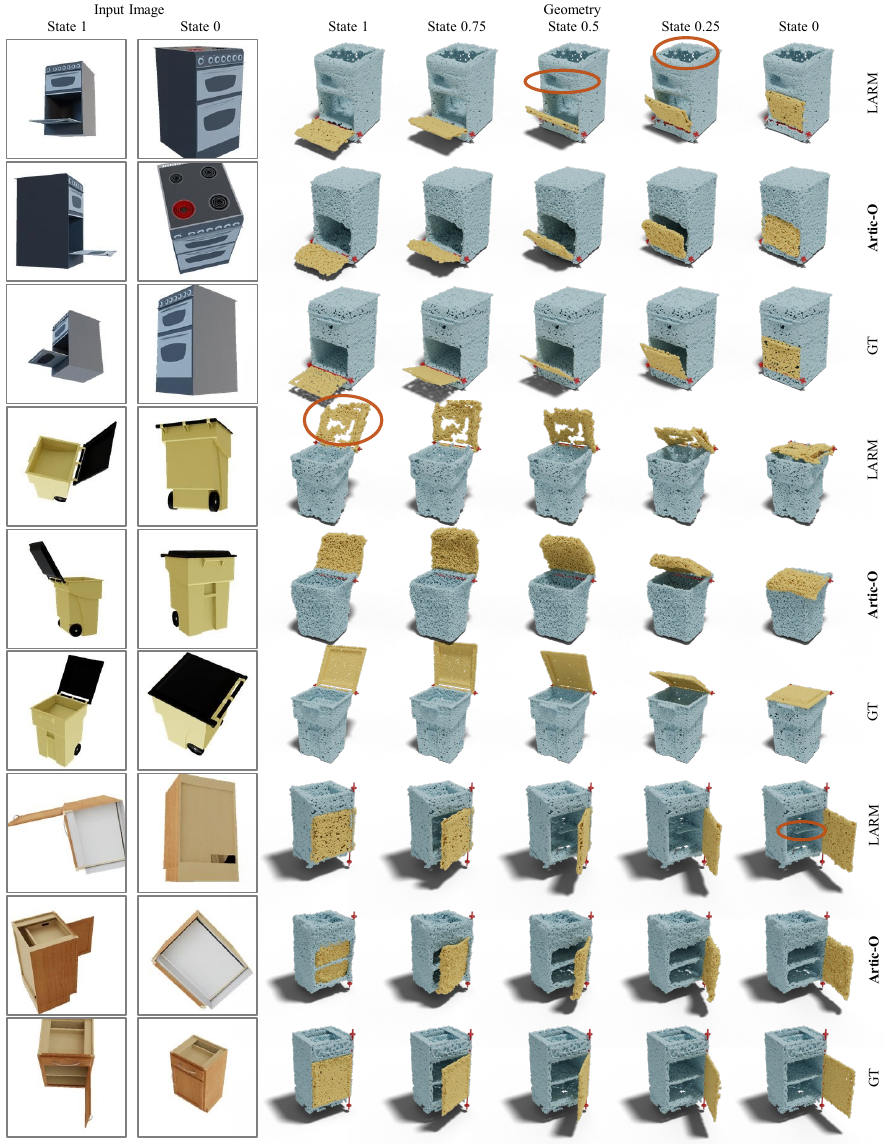}
    \caption{
Qualitative comparison between Artic-O and LARM~\cite{yuan2025larm}.
Both methods take sparse two-state images as input and reconstruct the articulated object across intermediate motion states.
Red circles highlight representative regions where LARM produces incomplete or inconsistent geometry, such as missing top surfaces, noisy lids, and broken interior structures.
    }
    \Description{Qualitative comparison between Artic-O and LARM}
    \label{fig:extra_2}
\end{figure*}

Fig.~\ref{fig:visual_compare} and Fig.~\ref{fig:extra_2} provide qualitative comparisons between LARM, \methodname, and the ground truth.
Compared with LARM, \methodname reconstructs more complete and coherent object shapes, especially around articulated parts and sparsely observed regions.
The reconstructed point clouds are visually closer to the ground truth, consistent with the stronger CD and F-score reported in Table~\ref{tab:main}.

\begin{figure}[t]
    \centering
    \includegraphics[width=0.9\linewidth]{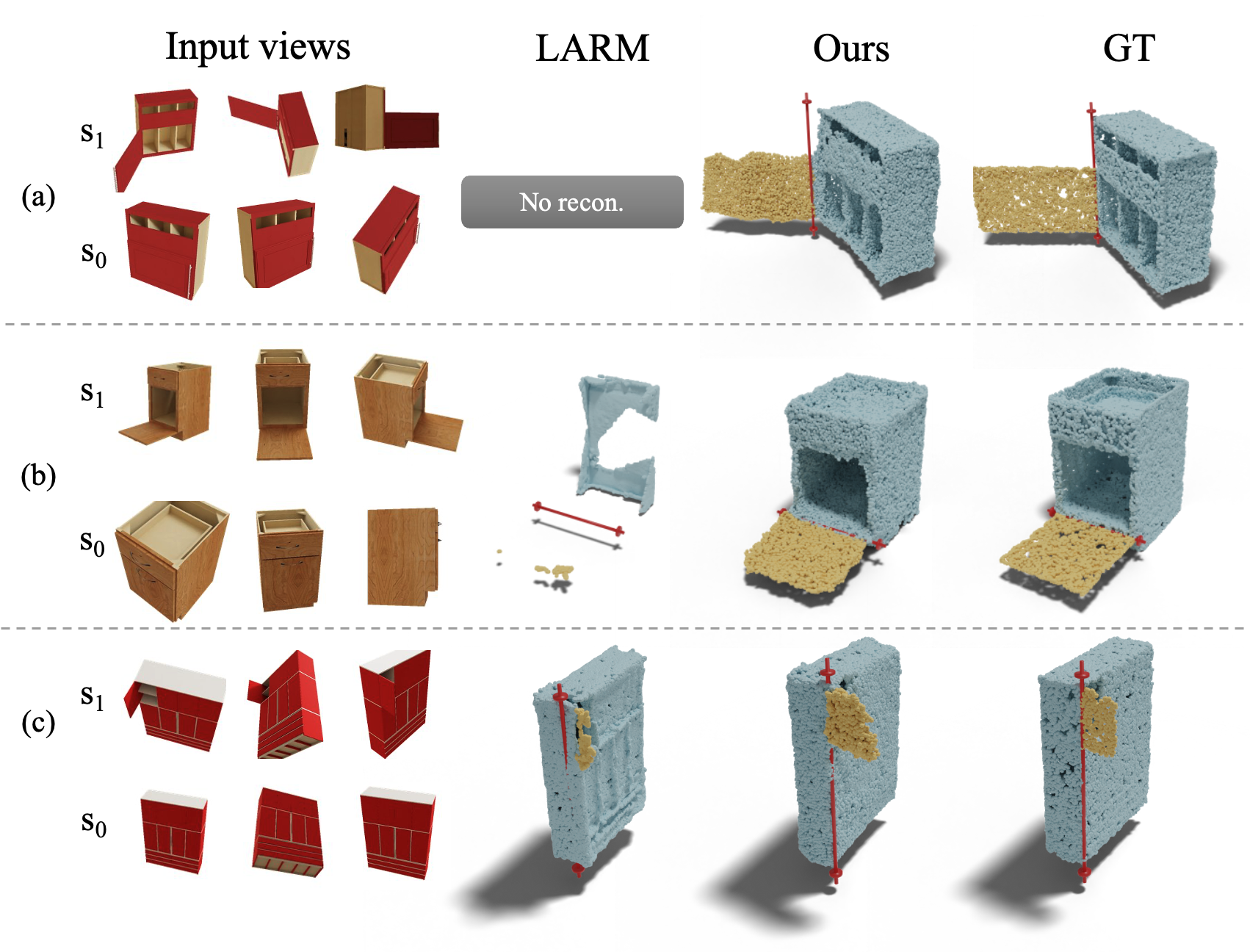}
    \caption{
    \textbf{Representative failure modes of LARM's multi-stage pipeline.}
    We visualize representative cases where LARM fails in reconstruction or articulation estimation, compared with \methodname and the ground truth.
    }
    \Description{Representative failure modes of LARM's multi-stage pipeline.}
    \label{fig:larm_fail}
\end{figure}

We further analyze the robustness of LARM's multi-stage inference pipeline under its released configuration and show representative failure cases in Fig.~\ref{fig:larm_fail}.
In (a), fixed depth truncation during TSDF fusion removes most of the useful depth observations, preventing a valid mesh reconstruction.
In (b), the same truncation removes approximately $99\%$ of the movable-part depth, yielding severely incomplete geometry despite a valid articulation estimate.
In (c), geometry reconstruction succeeds, but correspondence-based joint fitting yields inaccurate articulation.
Case (c) further illustrates a limitation of reconstruction-first approaches: although plausible 3D shapes are recovered at multiple states, the correspondence-based fitting stage can remain a source of error for articulation.

We also profile inference time on a single H200 GPU with batch size one.
Our end-to-end pipeline takes $0.32$ seconds per object, including the image encoder, 25-step Euler integration, segmentation and articulation heads, and post-processing.
By contrast, LARM takes about $9$ minutes in our implementation, dominated by repeated novel-view synthesis and correspondence-based axis estimation.
This corresponds to an approximately $1680\times$ speedup on the profiled samples, demonstrating the efficiency advantage of direct feed-forward prediction.
We quantitatively evaluate the part-level reconstruction quality on our test set. Specifically, \methodname achieves an F-score of $84.0\%$ on movable-part reconstruction, compared with $81.4\%$ for LARM~\cite{yuan2025larm}. Additionally, we evaluate active-part segmentation, where \methodname achieves an mIoU of $0.86$. We also include further examination of generalization to unseen categories in Appendix~\ref{sec:app:ood} and the effect of input resolution in \ref{sec:app:resolution}.

\subsection{Ablation Study}
\label{sec:exp:ablation}
\begin{table}[t]
    \centering
    \caption{
    \textbf{Ablation study of \methodname.}
Each row changes one design choice from the full model and keeps all other settings fixed, including the training budget and evaluation protocol.
    }
    \label{tab:ablation}
    \resizebox{\linewidth}{!}{
    \begin{tabular}{l|cc|cccc}
        \toprule
        \multirow{2}{*}{\textbf{Variant}} & \multicolumn{2}{c|}{Geometry} & \multicolumn{4}{c}{Axis} \\
         & \textbf{CD} $\downarrow$ & \textbf{F1@0.05} $\uparrow$ & \textbf{Angle} $\uparrow$ & \textbf{Origin} $\uparrow$ & $\mathbf{M_r}$ $\uparrow$ & $\mathbf{M_d}$ $\uparrow$ \\
        \midrule

        Full model & 0.017 & 95.4\% & 96.0\% & 98.0\% & 93.6\% & 94.4\% \\
        \midrule
        Random-init FM decoder & 0.020 & 93.5\% & 94.4\% & 98.8\% & 94.0\% & 92.3\% \\
        Unfreeze FM decoder & 0.017 & 95.4\% & 93.2\% & 98.0\% & 92.7\% & 89.5\% \\
        \midrule
        w/o state embedding & 0.017 & 95.3\% & 92.3\% & 98.0\% & 92.7\% & 89.9\% \\
        Point-only segmentation & 0.019 & 93.4\% & 83.1\% & 98.0\% & 90.3\% & 83.9\% \\
        w/o part memory & 0.025 & 92.5\% & 76.2\% & 90.3\% & 80.7\% & 76.6\% \\
        \midrule
        w/o curriculum & 0.029 & 82.8\% & 82.3\% & 94.4\% & 89.1\% & 82.3\% \\
        w/o two-pass training & 0.022 & 92.8\% & 78.2\% & 89.1\% & 86.7\% & 81.1\% \\
        
        \bottomrule
    \end{tabular}
    }
    \Description{Ablation study of Artic-O.}
\end{table}

Table~\ref{tab:ablation} studies the contribution of the main design choices in \methodname.
Starting from the full model, each variant removes or relaxes one component while keeping the remaining architecture, training budget, and evaluation protocol unchanged.

\textit{(1) Frozen geometry prior.} Randomly initializing the shape decoder degrades geometry, increasing CD from $0.017$ to $0.020$ and reducing F1 by $1.9$ points.
In contrast, unfreezing the pretrained shape decoder preserves reconstruction quality but weakens articulation prediction, reducing articulation accuracy by $2.8$ points and $M_d$ by $4.9$ points.
This indicates that the pretrained FM decoder is not only useful for geometry reconstruction, but also provides a stable latent geometry space for downstream part and articulation reasoning; preserving it as a frozen prior yields the best overall performance.

\textit{(2) Image-grounded part reasoning.} Replacing the slot-based part-attention module with a point-only classifier reduces axis-angle success by $12.9$ points and $M_d$ by $10.5$ points, indicating that point features alone are insufficient for reliable global motion prediction.
Removing the image-grounded part memory further degrades both reconstruction and articulation, increasing CD by $47\%$ and reducing axis-angle success by $19.8$ points.
Thus, the part memory provides a useful bottleneck for injecting multi-view, multi-state visual evidence into part-level reasoning.

\textit{(3) Training strategy.} Removing the curriculum produces the largest geometry degradation, increasing CD by $72\%$ and reducing F1 by $12.6$ points.
Without two-pass training, articulation drops substantially, with axis-angle success decreasing by $17.7$ points and $M_d$ by $13.3$ points.
These results confirm that reconstruction and part-level prediction benefit from different training regimes: flow matching uses noisy trajectory samples, while segmentation and articulation require features with clearer part identity.

\textit{(4) State-aware encoding.} Removing the state embedding has a smaller but consistent effect, increasing CD by only $2\%$ while reducing axis-angle success and $M_d$ by $3.6$ and $4.4$ points.
This suggests that explicit state labels help associate observations across articulation states, but become partly redundant once image-grounded part attention and two-pass supervision are used.

\begin{figure}
    \centering
    \includegraphics[width=0.9\linewidth]{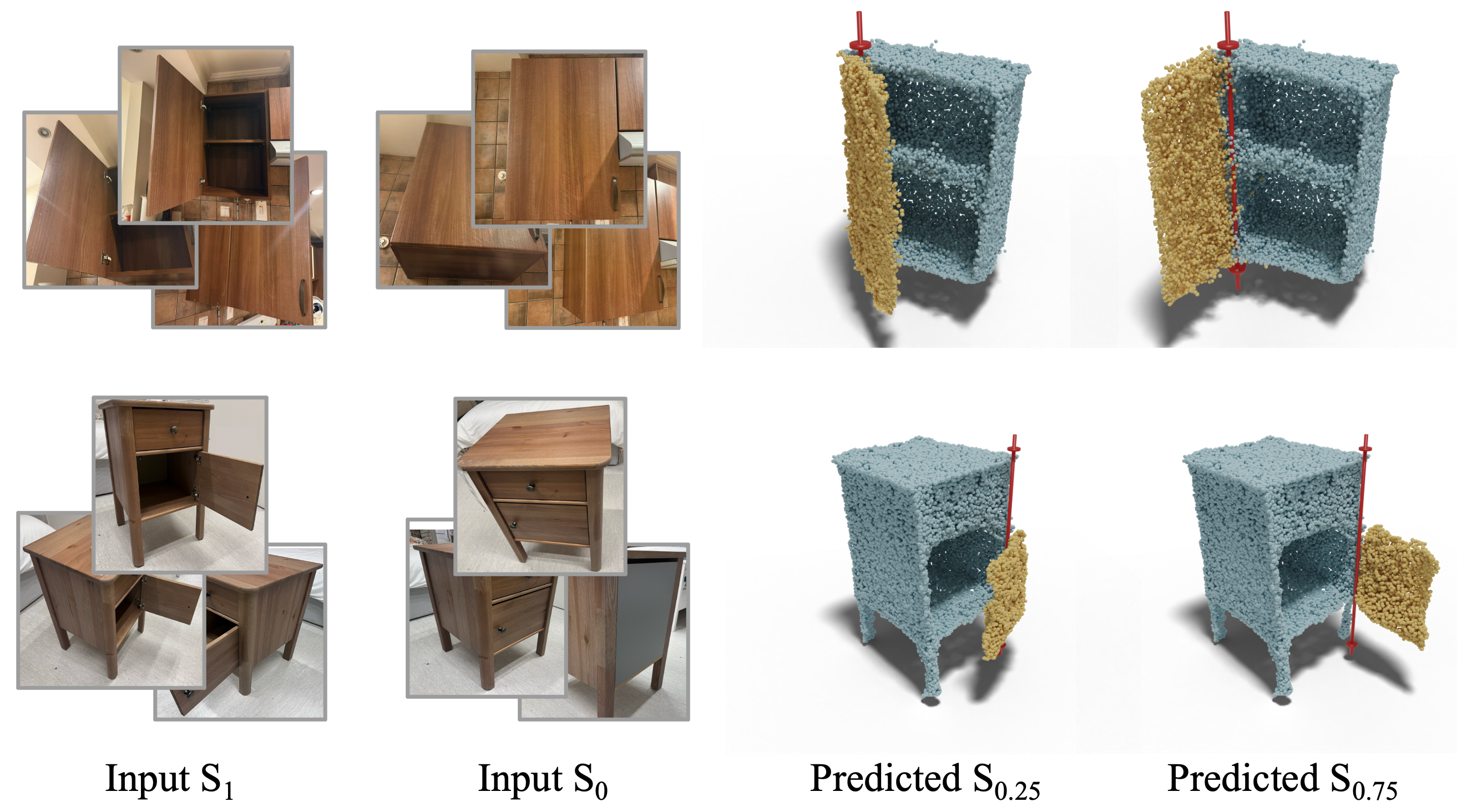}
    \caption{
\textbf{Real-world reconstruction from iPhone captures.}
 We show two real-world examples (one per row), with multi-view images at State~1 (fully open) and State~0 (fully closed) on the left, and reconstructions articulated to states 0.25 and 0.75 on the right.}
\Description{Real-world reconstruction from iPhone captures.}
\label{fig:realword}
\end{figure}

\subsection{Real-world Evaluation}
\label{sec:exp:realworld}

Beyond synthetic benchmarks, we qualitatively test \methodname on real-world images captured with a phone camera.
Following the same sparse two-state setting, we capture a few views of everyday articulated objects at two endpoint states and feed them directly to our model without test-time optimization.
As shown in Fig.~\ref{fig:realword}, \methodname reconstructs plausible geometry and predicts a meaningful active-part segmentation and articulation parameters.
This demonstrates the potential of our feed-forward pipeline to generalize beyond synthetic renderings toward real-world applications.

\begin{table}[t]
    \centering
    \caption{
    \textbf{Comparison of multi-part reconstruction.}
     Per-part inference spends $K$ forward passes for an object with $K$ movable parts; the one-pass variant emits all parts in a \textit{single} pass. 
    }
    \label{tab:multipart}
    \scalebox{0.8}{
    \begin{tabular}{p{4.3cm}|p{1.2cm}p{1.2cm}}
    \toprule
     Method & CD$\downarrow$ & F-Score$\uparrow$  \\
    \midrule
    LARM~\cite{yuan2025larm}  & 0.022 & 92.3\%  \\
    Artic-O, per-part (Ours) & \textbf{0.018} & \textbf{95.4\%} \\
    Artic-O, one-pass (Ours) & 0.023 & 91.0\% \\
    \bottomrule
    \end{tabular}
    }
    \Description{Comparison of multi-part reconstruction.}
\end{table}

\subsection{Extension to Multi-Part Inference}
\label{sec:exp:multipart}

We extend \methodname to multi-part articulated objects reconstruction and articulation using two schemes, described as followed.

In the first scheme, following LARM~\cite{yuan2025larm}, we process each movable part independently using our original single-part model without retraining. This requires $K$ forward passes for an object with $K$ movable parts.
In the second scheme, we instead extend \methodname from a single active slot to $K$ active slots, allowing all movable parts to be predicted jointly in a single forward pass.

We evaluate under the same multi-part protocol as LARM.
For each object, all joint states are sampled simultaneously to generate five articulated instances.
Metrics are first averaged over the five instances and then averaged across objects.
As shown in Table~\ref{tab:multipart}, the per-part scheme improves CD from $0.022$ to $0.018$ and F-score from $92.3\%$ to $95.4\%$ over LARM.
More importantly, the one-pass variant achieves a CD of $0.023$ and an F-score of $91.0\%$ while predicting all movable parts in a single forward pass, demonstrating that our slot-based formulation can be naturally extended to multi-part inference.
\section{Discussion and Conclusion}
We introduced \methodname, an end-to-end feed-forward framework for reconstructing articulated objects from sparse multi-state images. 
By extending geometry-latent learning from static object reconstruction to articulated reconstruction, \methodname predicts complete geometry, active-part segmentation, and articulation without a decomposed reconstruction-then-articulation pipeline or test-time optimization. 
Experiments on PartNet-Mobility show that our method improves reconstruction quality over LARM, maintains accurate articulation estimation, and achieves substantially faster inference, and our ablations validate the effectiveness of our key design choices.

While \methodname achieves strong articulated object reconstruction, it has several limitations. It assumes known joint types and supports only revolute and prismatic joints. The multi-part extension also assumes a known number of movable parts and models joints independently, without kinematic dependencies. Moreover, \methodname outputs point clouds rather than meshes and is trained primarily on synthetic data due to limited paired real-world 2D--3D data. Future work will explore richer kinematics, mesh-based reconstruction, and improved real-world generalization.

\bibliographystyle{ACM-Reference-Format}
\bibliography{ref}

\begin{figure*}
    \centering
    \includegraphics[width=0.95\linewidth]{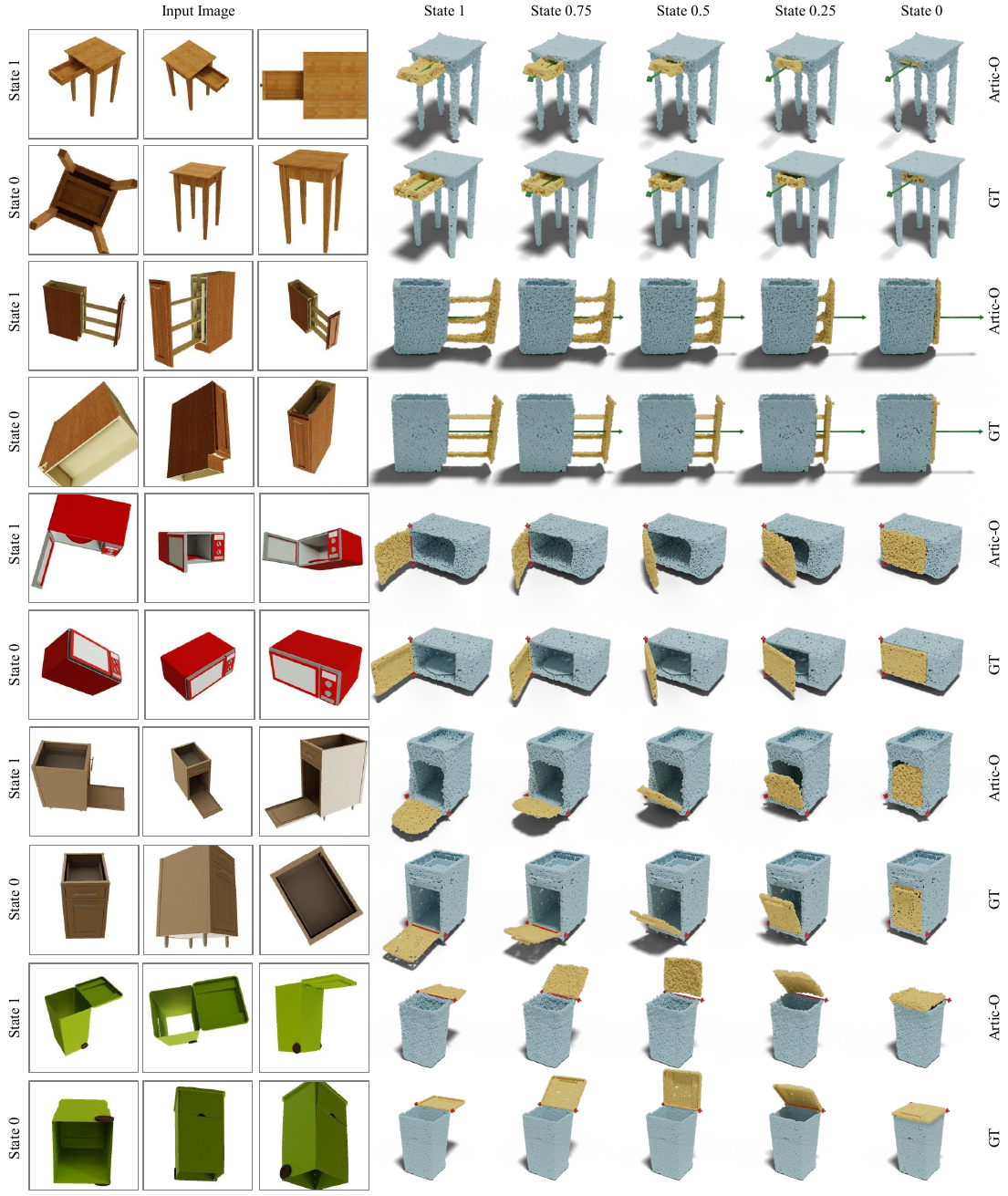}
    \caption{
Qualitative results of Artic-O.
We visualize the reconstructed object under five articulation states, where Artic-O directly predicts complete geometry, active-part segmentation, and articulation parameters from the input images.
    }
    \Description{Additional qualitative results of Artic-O.}
    \label{fig:extra_1}
\end{figure*}

\clearpage
\appendix

\section{More Experimental Results}

\subsection{Category Generalization}
\label{sec:app:ood}

We additionally evaluate category generalization on three PartNet-Mobility categories excluded from training: Oven, Washing Machine, and Suitcase. None of these categories appears in the training set.

Table~\ref{tab:unseen_ood} reports the F-scores on the unseen categories.
\methodname consistently outperforms LARM, indicating improved geometry reconstruction under category shift.
At the same time, we observe that generalization remains limited by the diversity and scale of available articulated training data, especially for categories with substantially different geometry or motion mechanisms.

\begin{table}[h]
    \centering
    \caption{
    \textbf{Category generalization to unseen categories.}
We report the F-scores$\uparrow$ for both Artic-O and Larm on three unseen categories. None of these categories appears during training.
Artic-O outperforms LARM, demonstrating improved generalization to unseen articulated object categories.
       }
    \label{tab:unseen_ood}
    \scalebox{0.88}{
    \begin{tabular}{p{3cm}|p{2cm}p{2cm}}
    \toprule
     Category & LARM~\cite{yuan2025larm} & Artic-O (Ours)  \\
    \midrule
    Oven  & 87.3\% & \textbf{88.9\%}  \\
    Wahing Machine & 83.2\% & \textbf{84.0\%} \\
    Suitcase & 88.7\% & \textbf{89.4\%} \\
    \bottomrule
    \end{tabular}
    }
\end{table}

\subsection{Discussion of Failure Cases}

\begin{figure}[h]
    \centering
    \includegraphics[width=0.95\linewidth]{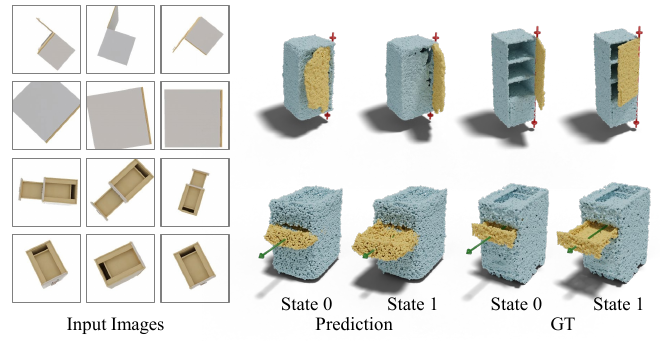}
    \caption{
    \textbf{Limitations.}
    Failure cases mainly occur when sparse input views insufficiently capture the movable part due to self-occlusion or limited visibility.
    These cases can lead to incomplete geometry, inaccurate active-part segmentation, or unstable articulation estimation.
    }
    \Description{Limitations.}
    \label{fig:failure_cases}
\end{figure}

We observe two main types of failure cases.
First, \methodname can struggle when test shapes or motion mechanisms deviate substantially from the training distribution.
This suggests that, like other feed-forward reconstruction approaches, its generalization ability is still bounded by the scale and diversity of available articulated-object training data.
Scaling training to broader object categories, articulation types, and motion patterns is a promising direction for improving robustness.

Second, as shown in Fig.~\ref{fig:failure_cases}, the sparse two-state input setting may not always provide sufficient visual evidence for reliably identifying the movable part.
This is especially challenging when the active component is weakly visible or self-occluded across the input views.
In such cases, the predicted active-part segmentation and articulation parameters can become unstable.
These limitations highlight the importance of both larger-scale articulated-object datasets and more informative sparse-view capture protocols for future work.

\subsection{Reconstruction Sharpness of Input Resolution}
\label{sec:app:resolution}
\begin{figure}[t]
    \centering
    \includegraphics[width=\linewidth]{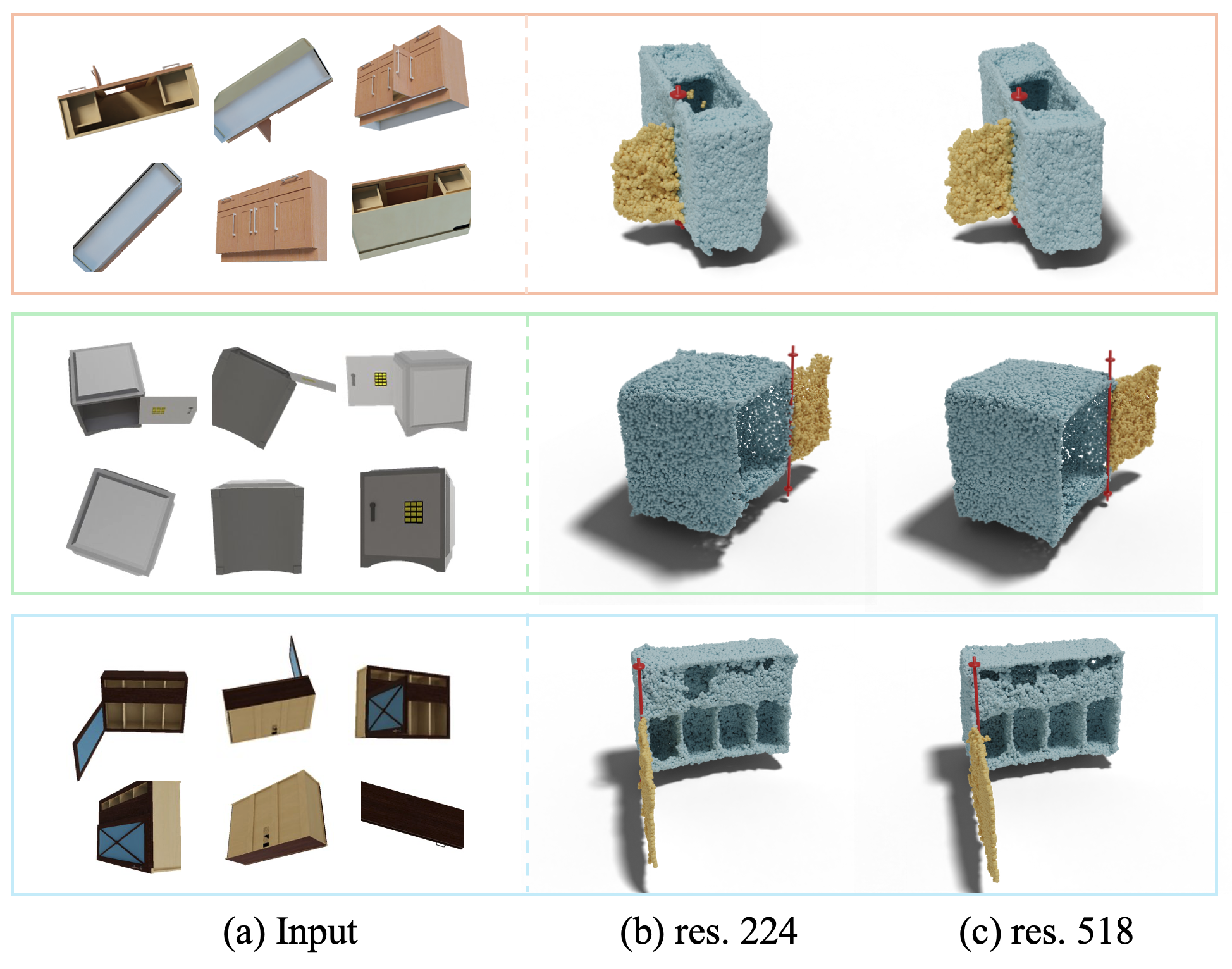}
    \caption{
    \textbf{Comparison of the Reconstruction Result with Different Resolutions.} 
    (a) Sparse multi-state input images. (b) Reconstructions using the default $224{\times}224$ input resolution. (c) Increasing the input resolution to $518{\times}518$ produces sharper local geometry and better preserves fine-scale structures, while retaining consistent overall geometry and articulation.
    }
    \Description{Comparison of the Reconstruction Result with Different Resolutions.}
    \label{fig:high_res}
\end{figure}

To examine whether resolution difference contributes to the relatively smoother geometry produced by \methodname, we additionally evaluate our model with $518{\times}518$ input images. Qualitative comparisons are shown in Fig.~\ref{fig:high_res}.

Increasing the input resolution improves the reconstruction of local geometric details across the examples. Compared with the $224{\times}224$ setting, the $518{\times}518$ results exhibit sharper part boundaries and better preserve thin or fine-scale structures, while maintaining consistent overall object geometry and articulation estimates. This comparison suggests that the reduced geometric sharpness observed in some of our main qualitative results is partly attributable to the lower-resolution image inputs. 

\end{document}